# AN IMPLEMENTATION OF COMPUTER GRAPHICS AS PREPRESS IMAGE ENHANCEMENT PROCESS

Jaswinder Singh Dilawari
Research Scholar, Dept of Computer Sci
Pacific University, Udaipur
Rajasthan (India)
dilawari.jaswinder@gmail.com

Dr Ravinder Khanna
Principal
Sachdeva Engineering College for Girls, Mohali
Punjab (India)
ravikh_2006@yahoo.com

## ABSTRACT

The production of a printed product involves three stages: prepress, the printing process (press) itself, and finishing (post press). There are various types of equipments (printers, scanners) and various qualities image are present in the market. These give different color rendering each time during reproduction. So, a color key tool has been developed keeping Color Management Scheme (CMS) in mind so that during reproduction no color rendering takes place irrespective of use of any device and resolution level has also been improved.

## 1. INTRODUCTION

Reproduction of high quality image in prepress step before printing is very crucial. Today various types of images and texts are produced by various people, so in a printing shop work has to divide which includes all three steps – prepress, printing process and finishing. Below given figure shows this work division. A printing press can get image from various types of equipments.

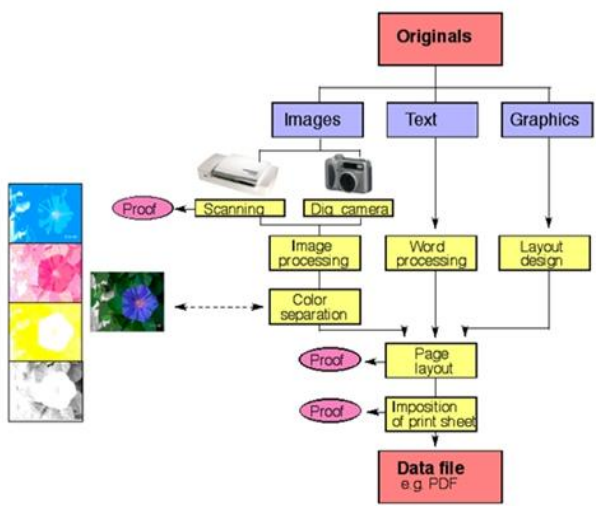

**Figure 1: The purpose of the page layout is to create a digital page from individual elements**

Computer graphics is used to reproduce the image in prepress process. In the process of reproduction improvement in resolution and quality of image is also a matter of concern. To achieve a print result with predictable color is thus complicated. A great help is "color management which attempts to make color more predictable within the limitations of the devices in use. Color management translates color between devices using a device-independent profile connection space and standard profiles for each device. A profile characterizes a device´s color reproduction capabilities. The color units (for example scanner, display, printer) are characterized in a common general format, ICC (International Color Consortium). Through the ICC-format, ICC-profile, various colors and hues can be interpreted in a similar fashion regardless of the platform and application (computer type, monitor model, system construction and pre-press programs), illustrated in Figures 2 and 3.

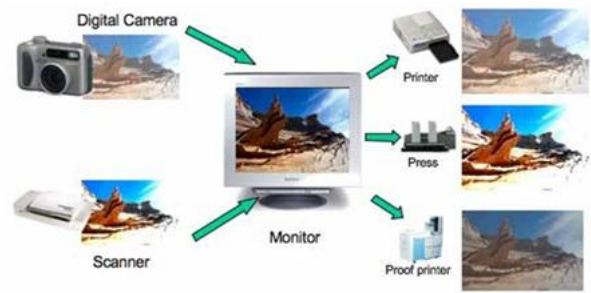

**Figure2: An example of image reproduction without color management**

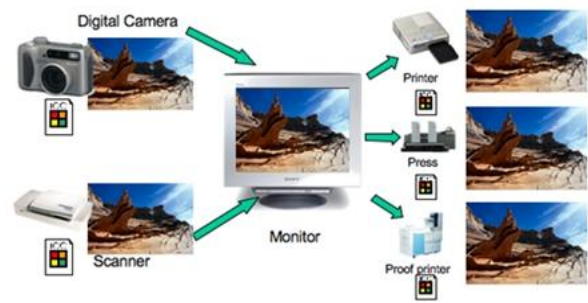

**Figure3: An example of image reproduction with color management**

*CMS:* Color management scheme (CMS) is a collection of color management software which produces the same image on monitor whatever be device is used. To obtain accurate and consistent color profile in devices four "Cs" are required - consistency, calibration, characterization and conversion.

*ICC:* International Color Consortium was formed in 1993 to seek to establish specifications and guidelines for the manufacturers and developers of software, equipment, and producers in terms of color management systems. ICC made an open profile format named "The ICC Profile Format" which can be easily used by every vendor. By defining a format which allowed consumers to mix and match profiles created by different vendors, the ICC standardized the concept of profile-based color management

*ICC Profile:* ICC Profile is set of data which describes color management of printers, faxes, scanners, digital cameras etc. The profile contains the information regarding any device in numerical data form. Numerical data is set of matrices which a color management module (CMM) uses to convert a device's color management to a common color space.

*CMM:* Color Management Module acts as intermediate between color space and PCS and PCS to any destination device. It basically converts the RGB or CMYK values using the color data in the profiles.

*PCS:* Profile Connection Space standard color space is the interface which provides an unambiguous connection between the input and output profiles, as illustrated in Figure 4. It is the virtual destination for input transforms and the virtual source for output transforms. To obtain consistent and predictable result input and output transform based on the same PCS definition can be paired by color management module (CMM) even at run time.

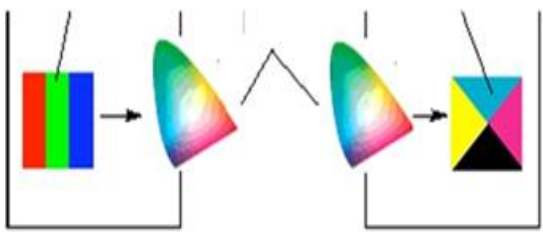

**Figure 4: A profile contains two sets of values, RGB or CMYK device control values, and the corresponding CIE XYZ or CIE. The values are converted from source color space to the PCS and from the PCS to any destination space.**

## 2. IMAGE UNDERSTANDING

Image reproduction requires that image quality should be enhanced. For those areas of an image has to be chosen which contains less information about image. Those areas are used for image processing. In order to retain important details in an image, the tone compression needs to be correctly controlled, but when this is carried out manually, the emphasis is on the tone area which one wants to retain to the greatest extent, i.e. to the area to be preferentially viewed in the image (for example an advertisement and its specific image).

*Images and Image categories:* The main area of interest in an image is where maximum information about image is stored. On the basis of this choice in literature image is categorized as: high-key, normal key, low-key (Field, 1990), gray balance and tertiary color images. If a photographer place the object in high tone area then it is called high key and if place in the shadow area of image, it is called low key

*Histogram of Image:* Histograms are the key to understanding digital images, see example of a histogram in Figure 5. In this Figure the histogram shows how the 256 possible levels of brightness are distributed in the image. The histogram displays the tonal distribution of the pixels in the image based on their level of brightness, on the x-axis from dark (0) to light (255). The y-axis represents the total number of pixels in the image of each level of brightness. If the histogram has the peaks concentrated towards the side of the graph, this is a "low-key" image. It can also mean an under-exposed image. If the peaks are concentrated towards the right-hand side, the image is "high-key", Figure6.

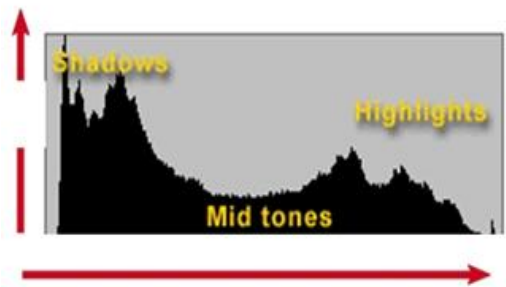

**Figure 5: The figure shows how to read a histogram**

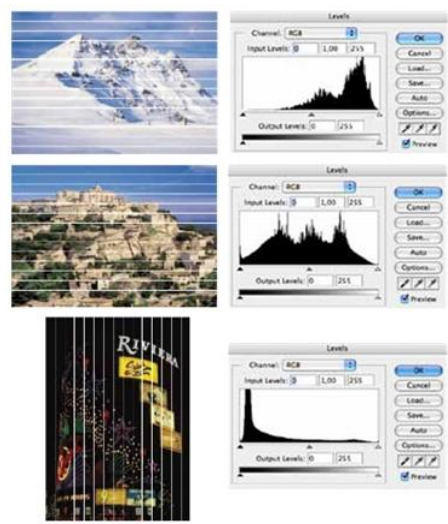

**Figure 6: Examples of a high-key image, a normal-key image and a low-key image.**

*Tonal Range and Tonal Compression:* In printing process various types of printing pages are used. For that toner level has to set as per the paper quality. Toner compression sometimes results in visible distortion areas by naked eyes. To be able to take the best possible advantage of the information in the original image, one should, during the scanning of the image, decide which areas of the image should be prioritized. Therefore, it is advisable to evaluate each image prior to scanning, and to decide which areas are of importance and which are not. Tonal reproduction is the most important aspect of color reproduction. It requires the high level compression so that best quality image can be printed. The compression should be uniform as shown in the figure 7.

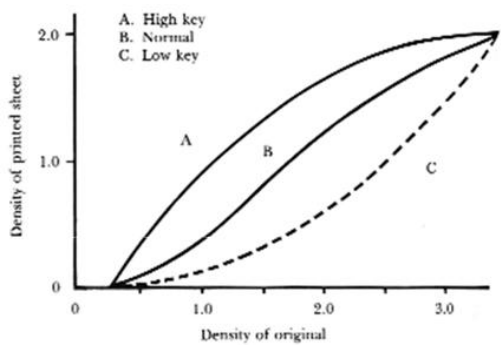

**Figure 7: Estimated tone reproduction curves for transparency reproduction, showing interest area emphasis for high-key, normal, and low-key photographs. A is a curve for the high-key image, B for the normal-key image and C for the low-key image**

## 3. STANDARDIZATION OF PRINTING PROCESS

Standardization of printing process leads to profitability and technical progress worldwide. For this The International Organization of Standardization (ISO) maintains many international standards, including those concerning color management, photography and printing. There are different" Technical Committees" (TCs) responsible for different fields. Concerning color management and printing, TC 130 is the most important committee. TC 130 maintains the following ISO standards:

*ISO 12640* Input data for characterization of 4-colour   process printing
*ISO 12642* SCID (standard color image data) images
*ISO 12647* process control for the manufacture of halftone color reparations, proof and production  prints
*ISO 13655* spectral measurement and colorimetric computation for graphic arts images
*ISO 15076* ICC color management
*ISO 15930* Prepress data exchange PDF/X

A work flow diagram of various standards representing s standards used at various levels is shown in figure 8.

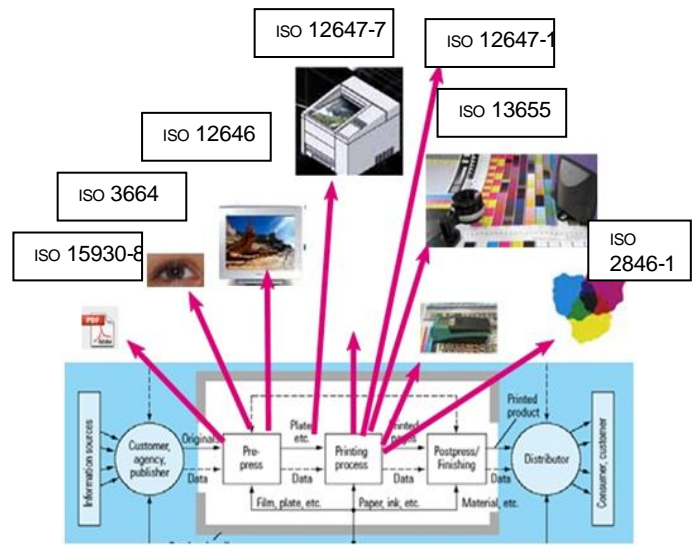

**Figure 8: Production flow, material and data flow for print media production. The majority of the graphical processes are currently supported by standards**

Our main focus will be only on standards used in prepress process because image reproduction quality will be increased at this level.

**ISO 15930** *Graphic technology - Prepress digital data  exchange* This ISO norm describes the requirements for PDF data being delivered to the printers.

**ISO 3664** *Graphic technology and photography - Viewing conditions* ISO 3664:2009 specifies viewing conditions for images on both reflective and transmit media, such as prints (both photographic and photomechanical) and transparencies, as well as images displayed in isolation on color monitors (ISO, 2009).

*ISO 3664:2009* applies in particular to:

•critical comparison between transparencies, reflection photographic or photomechanical prints and/or other objects or images;

• Appraisal of the tone reproduction and colorfulness of prints and transparencies at illumination levels similar to those for practical use, including routine inspection;

•critical appraisal of transparencies which are viewed by projection, for comparison with prints, objects or other reproductions;

•appraisal of images on color monitors which are  not viewed in comparison to any form of  hardcopy.

ISO 3664:2009 is not applicable to unprinted papers.

**ISO 12646** *Graphic technology -Displays for color proofing - Characteristics and viewing conditions*

ISO 12646:2008 specifies the minimum requirements for the properties of displays to be used for soft proofing of color images. Included are requirements for uniformity, convergence, refresh rate, display diagonal size, spatial resolution and glare of the screen surface. The dependence of colorimetric properties on the electrical drive signals and viewing direction, especially for flat panel displays, is also specified.

## 4. CONCLUSION

In printing process prepress process is very important from image quality point of view. Image is reproduced at this level. Use of computer graphics plays a vital role in enhancement of image. During image reproduction resolution of image should be enhanced keeping color management modules in mind. An area of image is selected such that if processing is done in that area then image quality doesn't reduce considerably. CMM defined by ICC helps in maintaining the quality of reproduced image irrespective of the device used for printing, scanning etc. By the use of computer graphics image should be enhanced in such a manner that after enhancement any considerable change can be noted.

## 5. REFRENCES